# Artificial Intelligence and Machine Learning to Predict and Improve Efficiency in Manufacturing Industry

I. EL HASSANI
Artificial Intelligence for Engineering Sciences Team
ENSAM-University My ISMAIL
Meknes, Morocco
i.elhassani@ensam.umi.ac.ma

C. EL MAZGUALDI
Artificial Intelligence for Engineering Sciences Team
ENSAM-University My ISMAIL
Meknes, Morocco
c.elmazgualdi@edu.umi.ac.ma

T. MASROUR
Artificial Intelligence for Engineering Sciences Team
ENSAM-University My ISMAIL
Meknes, Morocco
t.masrour@ensam.umi.ac.ma

*Abstract*—The overall equipment effectiveness (OEE) is a performance measurement metric widely used. Its calculation provides to the managers the possibility to identify the main losses that reduce the machine effectiveness and then take the necessary decisions in order to improve the situation. However, this calculation is done a-posteriori which is often too late. In the present research, we implemented different Machine Learning algorithms namely; Support vector machine, Optimized Support vector Machine (using Genetic Algorithm), Random Forest, XGBoost and Deep Learning to predict the estimate OEE value. The data used to train our models was provided by an automotive cable production industry. The results show that the Deep Learning and Random Forest are more accurate and present better performance for the prediction of the overall equipment effectiveness in our case study.

## I. Introduction

In order to be competitive in the era of globalization, manufacturing companies must improve and optimize their productivity while remaining flexible and responsive to customer demands. This fact led to the necessity of implementing a powerful performance measurement system which provides an unbiased process performance assessment. It consists of a set of Key Performance Indicators (KPIs) designed to allow managers to see the progress of their systems, it reflects the company's strategy and vision in term of objectives. So, it allows to follow both the targeted results and the actions (corrective or preventive) that achieve the objectives set. Key Performance Indicators are most often the result of a long chain of information gathering and aggregation, and they generally allow responsiveness and decisions are made more and more quickly.

The Overall Equipment Effectiveness (OEE) is a performance measurement metric wildly used. Its calculation provides to the managers the possibility to identify the main losses that reduce the machine effectiveness and then take the necessary decisions in order to improve the situation. Introduced by Seiichi Nakajima [1], OEE was a component of the Total Productive Manufacturing (TPM) concept. The losses can be divided into six categories ''Six Big Losses'' [2]. Fig. 1 shows these losses and some given examples of each one of them. One of the simplest ways to evaluate the The Overall Equipment Effectiveness (OEE), as in [1] and [3], is to apply the following formula:

$$OEE = A \times P \times Q \qquad (1)$$

Where:

$A$ = (Planned production time — Unplanned downtime) / Planned production time   (2)

$P$ = Actual amount of production / Planned amount of production   (3)

$Q$ = Actual amount of production — Non-accepted amount / Actual amount   (4)

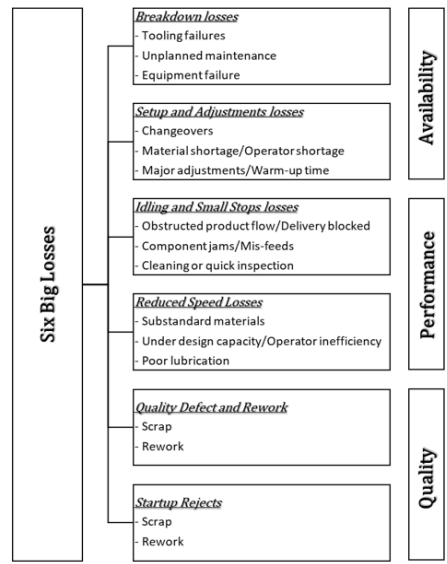

Fig. 1. Illustration of the Six Big Losses

The calculation of OEE, provides to the managers the possibility to monitor their process and identify the main losses that reduce the machine effectiveness [4]. However, this calculation is often done by the end of the production day and then it does not offer the possibility to act properly for improving the situation. The aim of this paper is to develop a Machine Learning based-model that predict in advance the estimate OEE value in order to allow for managers the possibility of evaluating the effectiveness of the equipment and therefore examining the inputs to the production process in order to identify and eliminate the relative losses. This will forms a decision support tool designed in the purpose of helping managers to prevent different kinds of losses and then act and react according to the given situation to improve and maximize the production effectiveness. In this research, four Machine Learning Algorithms are used, which are Support Vector Machine, Random Forest, Extreme Gradient Boosting, and Deep Neural Networks. The data used in this research is from the Cutting Department of an automotive wiring company.

## II. MACHINE LEARNING METHODS

Machine Learning algorithms are divided into three categories namely; Supervised Learning [5], Unsupervised Learning and Reinforcement Learning. For Supervised Learning, the problem is described by a vector X of input variables and a vector Y of output variables (Target) and the goal is to find a mapping function between X and Y as Y = f(X). In case we have only input variables X without any target variables, the problem is turned into an unsupervised learning. Finally, Reinforcement Learning is a different type of Machine Learning, where an agent learn how to behave and decide in an environment by performing actions and receiving rewards. In the present study, we will cover a variety of Supervised Machine Learning algorithms since we are handling labeled data.

### A. Support Vector Machine for Regression (SVR)

Support Vector Machine (SVM) [6-7] is a supervised machine learning algorithm mostly used in classification, but easily adopted for regression since it form a generalization of the classification problem, in which the model returns a continuous-valued prediction. SVR can be defined as an optimization problem that relied on defining a convex ε-insensitive loss function, then trying to minimize this function and find the flattest tube that contains most of the training instances. Based on the loss function and the geometrical properties of the tube, a multiobjective function is constructed which provides a unique solution to the convex optimization. The hyperplane is represented in terms of support vectors, which are training samples that lie outside the boundary of the tube. The figure below (Fig. 2) shows an example of one-dimensional linear regression function with – epsilon insensitive – band.

SVR technique requires certain parameters, mostly: the kernel and regularization parameter. Kernels are used when we are dealing with non-linearly separable data, they allow mapping the input data to a high-dimensional space in order to be linearly separable. The most used kernels are Radial Basis Function Kernel (RBF), Polynomial Kernel and Sigmoid Kernel [8]. Regularization parameter also known as the soft margin constant C: this hyper–parameter controls the trade-off between the slack variable penalty and the width of the margin. Finally, the complexity parameter, which is $\gamma$ for RBF, Poly and Sigmoid kernels and the degree for polynomial kernel, determine the flexibility of the SVM model in fitting the data while preventing Overfitting.

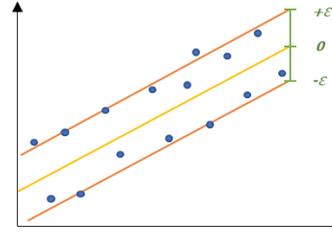

Fig. 2. One-dimensional SVR example

### B. Ensemble Learning Techniques

Ensemble methods are a collection of several base models working together on a single set of data in order to produce one optimal predictive model. Ensemble models are known for their higher accuracy results against conventional single-machine learning models [9]. Ensemble Learning techniques are classified into Bagging and Boosting.

Bagging, short for Bootstrap AGGregatING is a simple ensembling technique that consist of creating several subsamples of data from training sample chosen randomly with replacement. Then, several models are trained with the different subsets/bootstraps so that we end up with several predictions. Finally, these predictions are aggregated by using their weighted average, majority voting or simply their normal average. Random Forest is a bagging technique that we will cover in the present study. Boosting is a powerful ensembling technique in which the models are made sequentially. At each particular iteration, a new weak, base-learner model is added to be maximally correlated with the negative gradient of the loss function. The learning procedure of this technique is based on the logic that the subsequent predictors learn from the mistakes committed by the previous predictors.

Gradient Boosting Machine (GBM), Adaboost and eXtreme Gradient Boosting (XGBoost) are the most famous boosting algorithms, in our research we will cover XGBoost model.

*a)* Random Forest [9-10]: Random Forest is a supervised Machine Learning algorithm used for both prediction and classification problems. It is a bagging ensemble learning method based on Decision Tree algorithm. The random forest is a popular algorithm that can be applied to a wide range of prediction and classification problems and have few parameters to tune. It is forming by a multitude of decision trees, and the prediction is made by aggregating (majority vote for classification or averaging for regression) the predictions of each component tree as shown in Fig.3 (Y1, …,Yn are the predictions given by the different decision trees). It generally has much better predictive accuracy and show a good performance than a single decision tree, and it is recommended for its ability to deal with small and large sample sizes. The Random Forest is proper for high dimensional data modeling because it can deal with categorical, binary and continuous data and it can handle easily the problem of missing values.

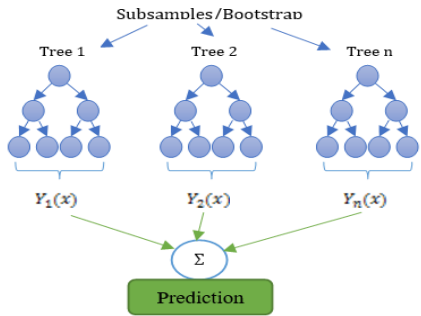

Fig. 3.  Bagging/Random Forest Example

*b)* eXtreme Gradient Boosting: XGBoost (eXtreme Gradient Boosting) is an optimized library designed for implementing machine learning algorithms under the Gradient Boosting framework [11]. It is a boosting ensemble learning method based on Decision Tree algorithm. By imposing regularization and providing parallel tree boosting, XGBoost is able to exploit more computational power and get more accurate prediction while remaining fast and scalable.

### C. Deep Learning Neural Network

The core difference between Deep Learning and Neural Network is that Deep Learning is a deep neural network. In other words, a Deep Learning model is created by a multilayers architecture of neural nets and is taking advantage of the huge amount of accessible data nowadays. Although its architecture is more complex and it use more and more data for training, Deep Learning remain more performing than any other Machine Learning algorithm. Furthermore, its performance keep increasing while increasing the number of data.

The deep learning algorithm consist of feed-forward and back-propagation process [12]. While the feed-forward process, the model associate weights from the input layer toward hidden units and the output layer. After every iteration through the data set, the weights are adjusted toward a backward propagation using Gradient Descent to maximize the correlation between the output and the residual error of the model. The final weights vector of a successfully trained neural network represents its knowledge about the problem [13].

## III. PROPOSED PREDICTIVE MODELS

In our study, we attempt to cover the most popular and famous Machine Learning algorithms. We first tried different linear regression models namely; Multiple Linear Regression, Polynomial Regression, Support Vector Regression (SVR) and Decision Tree Regression. Then, we kept SVR because it shows better accuracy within the regression models. The SVR algorithm is implemented under two configurations, the first model is built using the parameters concluded by trial and error. However, the second model used the hyper-parameters resulted by the Genetic algorithm. Secondly, we manipulated ensemble-learning techniques with tree-based models. We tested Random forest for Bagging Ensemble and XGBoost for Boosting Ensemble. Finally, we built a Deep Learning predictor in order to come up with a robust model with a high level of performance and especially able to deal with the complexity of our data.

### A. Support Vector Regression

As we discussed in the previous section, we chose Support Vector Machine for Regression (SVR) based on its accuracy. Furthermore, SVR algorithms are depict by their adaptability, flexibility to work with linear and non-linear problems, and mainly they are non-biased by outliers. Our SVR model is implemented under the architecture below: Kernel: Radial Basis Function (RBF), $C = 1$ and gamma = 0.15. In the second part of the SVR method, we developed a Genetic Algorithm (GA) for hyper-parameters tuning, especially C and gamma. As a result, we obtain the two optimized values, which are: $C = 403.47$ and gamma = 0.83. Then we built our second SVR model with the given parameters.

### B. Random Forest

Besides its accuracy and interpretability, Random Forest (RF) is also an appropriate model for high dimensional data, which make it suitable for our case study. Our RF model is implemented using 120 trees, the number of trees was decided by trial and error using mean absolute error (MAE) as a metric.

### C. eXtreme Gradient Boosting (XGBoost)

XGBoost is an ensemble tree-based model, which flow the principle of gradient boosting just as Gradient Boosting Machine (GBM) and Adaboost. However, XGBoost has more customizable parameters that allows it a better flexibility. Additionally, XGBoost use more regularized model formalization to control over-fitting, which gives it better performance. All of the above reasons combined with the computational power of the XGBoost algorithm, were sufficient for us to choose it over the other boosting techniques. Our XGBoost model is developed using the following hyper-parameters: Number of tree boosted: 130; Learning rate (eta) = 0.2; Maximum depth= 3.

*D. Deep Neural Networks*

Our Deep Learning (DL) model is implemented using an input layer of 7 neurons which represent our input features and two hidden layers. To estimate the number of neurons in each hidden layer, we checked different methods and heuristics [14-15-16] and finally we adopted Huang's network architecture for two hidden layers feedorward network (TLFN) [17]. It consists, in our case study, of 125 neurons in the first layer and 25 in the second layer. It remains to be noted that an architecture of 28 nodes in each hidden layer give a comparable result. The output layer consist of one node given that we are predicted one final value (OEE). The architecture used in this research is represented in Fig. 4.

The architecture of our DL used in this section is formed by the following structures and parameters: Learning rate: 0.001; Epoch number: 3000; Number of neuron in hidden layer L1=125 and L2=25; Activation function: Rectifier Linear Unit; Optimizer: Adam optimization algorithm [18]; Loss function: Mean Absolute Error.

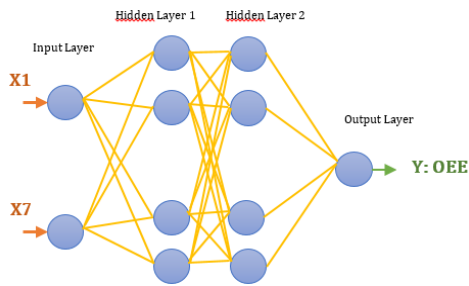

Fig. 4. Deep Neural Networks topology

## IV. EXPERIMENTS

*A. Industrial Case study*

The data used in this research was provided from the cutting and crimping department of an automotive wiring harness company. In terms of the production process of wire harness it consist of wire cutting and terminal crimping, then the wires and components are assembled on the pin-board to the desired specification and then bound together. Finally, electrical tests, and visual inspection are highly recommended from the quality side, an example of electrical wire is shown in Fig. 5 (left). In this study, the cutting machines are used as the telltale equipment. The cutting workshop is occupied by 22 similar machine which are designed to cut a range of wires cross-sections. The operations ensured by this equipment are as follows: After any required preparation, the first step is to cut the wires to the desired length. A special machine may also print the wires on during the cutting process. After this, the ends of the wires are then stripped to expose the core of the wires, usually metallic, Fig. 5 (middle) shows an example of a stripped wire. The next step is to fit the wires with the required terminals and/or connector hous-ings, which come in many different sizes and specifications. This step is known as crimping. Every product is a specific product which has its own specification.

The main specifications of a new product are as follows: Cable Class (Single wire, …); Wire cross section; Wire length; Stripping Length (Right and Left); Number of terminals (0, 1, 2); Terminal identification; Number of Seals (0, 1, 2); Print (Yes/No); Print Identification. Fig. 5 (right), represents an example of electrical wire with two different terminals.

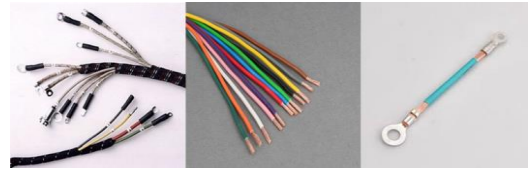

Fig. 5. Electric Wire Harness (left). Cut and stripped Wire (middle), Wire with two different terminals (right)

A production order is defined by a given quantity of one product and is launched by the planning department and then they are dispatched automatically to the different machines. One machine could produce a set of orders per day per shift.

*B. Data Preparation*

We evaluated our model on a dataset that we prepare based on the extracted data from the internal system of the company. The raw data was given into two tables, the first contains the OEE values per shift per machine and the second shows the different orders with the associate specifications done by every single machine over the days.

The data preparation process consist of the following stages [19]. 1) Data selection: which means selecting appropriate data from the raw data (the given two files). 2) Data preprocessing: the data preprocessing step consist of cleaning and removing fixing or missing data and then transforming the data into a structured form suitable to work with. Sampling the data is also a data-preprocessing step that was accomplished in this level. 3) Data transferring: it consist of features scaling that was highly recommended in our case -because our input variables did not has the same scale- in order to avoid the big-scale variable(s) domination. Finally, for augmenting our dataset, it was recommended to consider that the whole data was provided by one machine since the machines are similar.

In the first section of this paper, we presented the tree core parameters that has an impact on the OEE value. These parameters are the availability, the performance and the quality. However, the yields of all the machines in every observation of our dataset are 100%, and then the quality rate will not be taken into account. Thereby, we end up with a set of 12 variables that affect the two retained components of the OEE, which are performance and availability. By applying features selection approach, and choos-

ing the more correlated features to our target, our final set of input variables is formed by:
- Setups: This variable represent the sum of the all setup values. As the setup for every order is known before starting the production, this variable can be used as an input for our predictors.
- Breakdown: this variable is substitute by planned and unplanned maintenance. In our study this will concerns only planned maintenance, since unplanned maintenance cannot be known in advance.
- Number of orders: This variable act out how many orders every single machine has done per day per shift.
- Mean Wire Length: represent the mean of overall wire length values produced by the same machine in one shift.
- Number of Terminals: represent the total number of terminals across all orders for the same machine in one shift.
- Number of Seals: represent the total number of seals across all orders for the same machine in one shift.
- Target: this variable explain the target designed for each order regarding its difficulty and other parameters.

### C. Exploratory Data analysis (EDA)

Before starting our Machine Learning Modeling, a data exploration approach was needed [20]. We tried to detect the presence of outliers in our data so to drop them or even to remove them, this step was performed by graphical techniques. Using the distribution plot Fig. 6 (top-left) and the box plot Fig. 6 (top-right), we were able to detect the existent outliers.

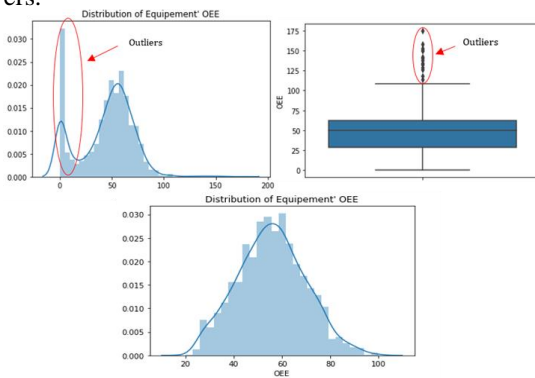

Fig. 6. Distribution plot of raw data (top-left). Box plot of raw data (top-right). Distribution plot of retained data (bottom)

By analyzing our data set, most of the outliers are meaningless aberrations caused by errors (under and over values) committed when reporting the daily information on the system and the rest are due to some situations that rarely occur. Hence, we removed all of these outliers from our dataset. Fig. 6 (bottom) shows the distribution plot of our final (retained) data set.

## V. EMPIRICAL ILLUSTRATION

In this case study, the dataset consist of 7 features, 1917 observations and 1 target (OEE value). All the models except DL will be implemented, first using a simple data split and secondly using cross validation technique. The data was split into training (85% of the data) and test (15% of the data), then we will take 40 items from our test data to elaborate the plots for the visualization and comparison purpose. All the models used in this research are developed using Python 3.6.

### A. Support vector regression results

As mentioned before, we implemented three SVR models. SVR with simple data split, SVR with cross validation (SVRCV) and optimized SVR with cross validation using Genetic algorithm (SVRGA).

*a) SVR:* Fig. 7 (left) shows the results for the 40 items. The average error for the test dataset is 12.12%. In Fig. 7 (left), it is found that the prediction values of SVR have the same trend as the real values and also they are slightly close to the targets. However, this might imply that our model tend to over-fit the data and then it will not be reliable if we change the data.

*b) SVRCV:* Fig. 7 (midle) shows the results for the 40 items. The average error for the test dataset is 19.99%. As expected, the first SVR model was not reliable enough and the error was not representative of the model accuracy. In Fig. 7 (midle), it is found that the prediction values of SVR using cross validation only have small variation. There is no obvious change for the 40 items of validation data.

*c) SVRGA:* In order to improve the previous model, we developed a Genetic algorithm for hyper-parameters tuning (C and gamma). Then the issued values of C and gamma was used for our SVRGA model which provides us a performed result. Fig. 7 (right) shows the results for the 40 items with an average error of 14.26%. The predicted values are close to the real ones and have more or less the same trend.

### B. Random Forest results

In the study, we implemented two RF models, a first model with simple split data and a second one with cross validation.

*a) RF:* The first RF model is built with the elementary split data. Fig. 8 (top-left) shows the results for the 40 items with an average error of 13.59%. It is found the trend of the prevision is almost the same as the target trend, also the predicted OEE values are too close from the real ones. But like we discussed for the SVR method, the simple split does not allows a functional model on unseen data.

*b) RFCV:* In order to improve the effectiveness of our model, we implemented the Random Forest algorithm using cross validation split data. Fig. 8 (bottom-left) shows the results for the 40 items. The average error for the test dataset (15% of data) is about 12.89%. In Fig. 8 (bottom-left), it is found that the prediction values of RFCV only have small variation compared to the results given without cross validation. Unlike for SVR, the RFCV kept its accuracy higher (better accuracy compared to RF) while performing a better fit to the data, it shows the robustess and the reliability of the model.

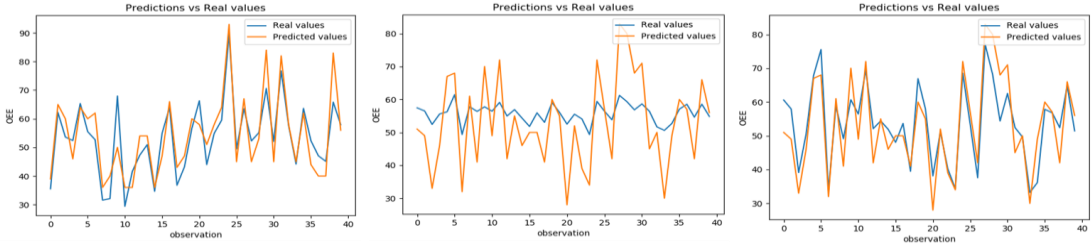

Fig. 7. SVR results (left), SVRCV results (midle), SVRGA (right)

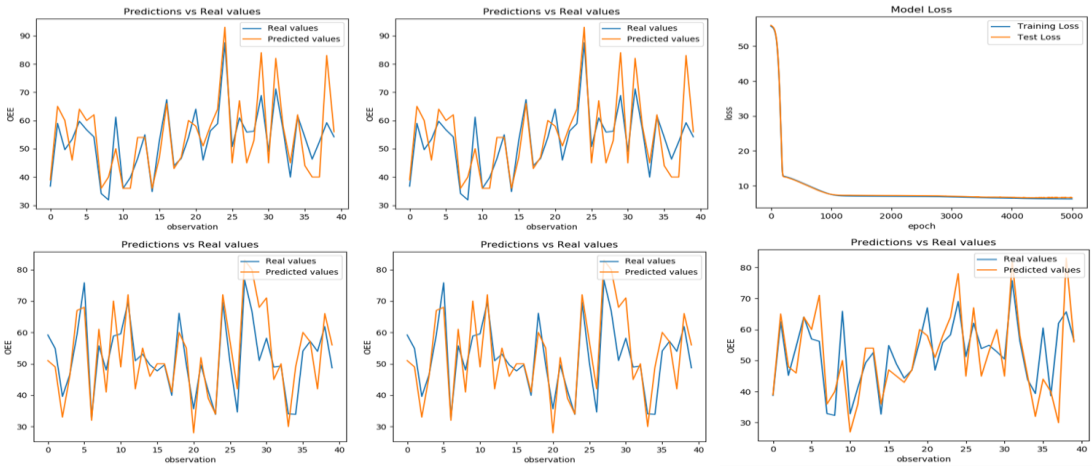

Fig. 8. RF results (top-left), RFCV results (bottom-left). XGB results (top-middle), XGB results (bottom-middle). DL loss curve (top-right), DL results (bottom-right)

## C. eXtreme Gradient Boosting results

The XGBoost algorithm is implemented under two configurations, a first model with simple split data (XGB) and a second one with cross validation (XGBCV).

*a) XGB:* The XGB model is built with the elementary split data. Fig. 8 (top-middle) shows the results for the 40 items. The average error for the test dataset is about 12.59%. In Fig. 8 (top-middle), it is found that the prediction values of XGB have the same trend as the real values and also they are slightly close to the targets. Yet, the same remark as for SVR and RF still valid.

*b) XGBCV:* In order to improve the model performance, we implemented it with cross validation split data. Fig. 8 (bottom-middle) shows the results for the 40 items of validation data, with an average error of 15.37%. In Fig. 8 (bottom-middle), it is found that the prediction values of XGBCV have a good trend regarding the real values. Although the average error has slightly increased, but like RFCV, XGBCV remain accurate, more performed and reliable on unseen date.

## D. Deep Learning results

In the case study, the Deep Learning (DL) model implemented consist of two hidden layers with 125 neurons in the first layer and 25 neurons in the second layer. The architecture of the model was described in the section 3. Our model is built without cross validation, because of the cost associated with training K different models. Instead of doing cross validation, we used an elementary split of the data consisting of 85% of data for training and the remained 15% for testing. Fig. 8 (top-right) shows the loss curve for both training and test sets. It is found that the two losses have the same decreased trend, which reflects the stability and the performance of the model. Fig. 8 (bottom-right) shows the results for the 40 items with an average error of 11.76%. In Fig. 8 (bottom-right), it is found that the prediction values of this model have a good trend regarding the real values. In addition, it present a good fit to the data compared to the others models.

## VI. MODELS COMPARISON & DISCUSSION

### A. Model Accuracy

A quick comparison of our models based on their Mean Absolute Errors (MAE) and Mean Absolute Percentage Errors (MAPE) (see TABLE I.) shows that all the models without cross validation have good performance, but they are highly exposed to be biased and unreliable on future unseen data. DL and RF are as we discussed their performance in the previous section. We can conclude from the table that DL, SVRGA, RFCV and XGBCV show a good performance. Moreover, DL is more suitable to deal with the complexity of the data compared to the other models.to deal with the complexity of the data compared to the other models.

TABLE I. DEVELOPPED MODELS AND THEIR PERFORMANCE INDICES

| Predictive models | Metrics | |
|---|---|---|
| | Mean absolute error | Mean absolute percentage error (%) |
| SVR | 6.16 | 12.12 |
| SVRCV | 9.82 | 19.99 |
| SVRGA | 7.05 | 14.26 |
| RF | 6.83 | 13.59 |
| RFCV | 7.39 | 12.89 |
| XGB | 6.43 | 12.59 |
| XGBCV | 7.55 | 15.37 |
| DL | 6.27 | 11.76 |

Another evaluation is performed using statistical tests to compare means, namely with ANOVA test, and Standard Deviations Test of the Absolute Value Error respectively named : SVRCVGA-e; XGBCV-e; SVRC V-e; RF-e; RFCV-e; SVR-e; XGB-e; DNN-e.

### B. ANOVA test

We performed the ANOVA test with the following hypotheses: The null hypothesis is all means are equal. We take Significance level α = 0.05. The P-value is less than α. Therefore, the null hypothesis is not valid, and we can conclude that there is a significant difference between the means of model errors. The models: DNN; XGB; SVR; RFCV and RF have lower means compared to SVRCV; XGBCV and SVRCVGA. In Fig. 9, we grouped models using the Tukey method with 95% Confidence. Fig. 10 shows the existence of two significantly different model groups A and B.

### C. Standard Deviations Test

We performed the Standard Deviations Test with significance level α = 0.05. The P-value is less than α. Therefore, the null hypothesis is not valid, and we can conclude that there is a significant difference between the dispersion of model errors. The models: DNN; XGB; SVR; RFCV and RF have lower dispersions compared to SVRCV; XGBCV and SVRCVGA as shown in Fig. 11.

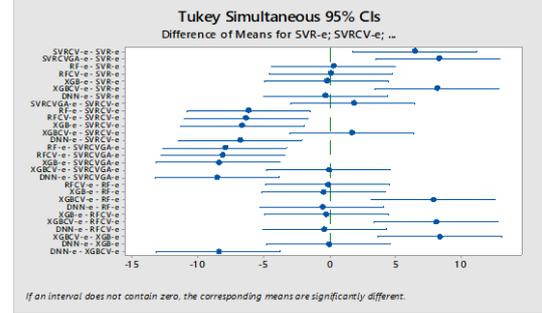

Fig. 9. Tukey Simultaneous 95% CI's

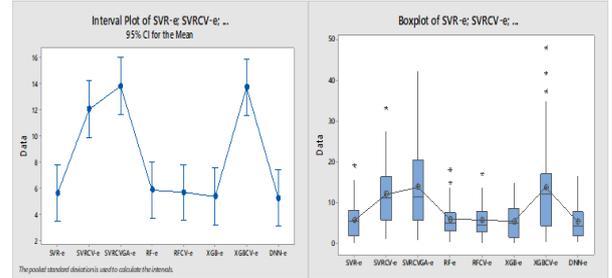

Fig. 10. Interval Plot of models' error (left), Boxplot of models' error (right)

### D. Conclusion of Comparison

If we take into consideration the fact that some models can be subject to over fitting, namely: XGB; SVR; and RF, we can then only retain 5 models for comparison, namely: DNN; RFCV; SVRCV; XGBCV and SVRCVGA.

We can conclude that DNN and RFCV have significant lower means and dispersions compared to SVRCV; XGBCV and SVRCVGA.

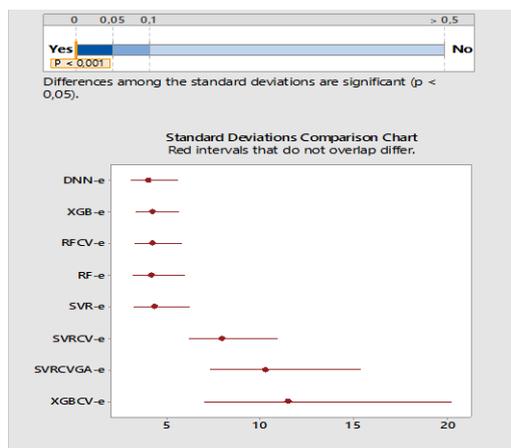

Fig. 11. Standard Deviations Comparison

## VII. Conclusion & perspectives

This paper presented the use of several Machine Learning algorithms for the prediction of organization KPIs, the application was conducted on the OEE (overall equipment effectiveness) where we demonstrated how precise prediction could be done using artificial intelligence techniques. Based on a real dataset provided by the cutting department of an automotive wiring company, we experimented several machine learning algorithms to predict the OEE value. The result exhibit that Deep Learning and Random Forest with cross validation shows better reliability and performance dealing with given data. The OEE is only a study case that can be generalized to other KPIs.

It should also be noted that many points remain to be explored as future work. We can cite: - The validation of the prediction in the real-time case, - The combination of optimization techniques with the developed models to provide Decision Support Tools for managers, - The more detailed study of the utility of the hybrid approaches which associate the genetic algorithms with those of the machine learning.